\documentclass[runningheads]{llncs}
\usepackage{graphicx}
\usepackage{amsmath,amssymb}
\usepackage{color}

\usepackage{hyperref}
\usepackage{cite}
\usepackage{algorithm}
\usepackage{algorithmic}
\usepackage{subfigure}
\usepackage{epstopdf}
\usepackage{booktabs}
\usepackage{array}
\usepackage{multirow}
\usepackage{hhline}
\usepackage{makecell}
\usepackage{framed}
\usepackage{enumitem}
\usepackage{authblk}

\usepackage{epsfig}
\usepackage{graphicx}
\usepackage{amsmath}
\usepackage{amssymb}
\usepackage[british,UKenglish,USenglish,english,american]{babel}

\newcommand{\fref}[1]{Fig.~\ref{#1}}

\graphicspath{{images/}{images/restoration/}}

\begin{document}
% \linenumbers
%\pagestyle{headings}
%\mainmatter
%\def\ECCV18SubNumber{578}  % Insert your submission number here

\title{Decouple Learning for Parameterized Image Operators} % Replace with your title

\titlerunning{Decouple Learning for Parameterized Image Operators}
% Replace with a meaningful short version of your title

\author{Qingnan Fan$^{1,3}$\thanks{Equal Contribution}, Dongdong Chen$^{2}$$^{\star}$, Lu Yuan$^4$, Gang Hua$^4$,
Nenghai Yu$^{2}$, Baoquan Chen$^{5,1}$}

\authorrunning{Q. Fan, D. Chen, L. Yuan, G. Hua, N. Yu, B. Chen}

\institute{$^{1}$Shandong University, $^{2}$University of Science and Technology of China\\
 \email{fqnchina@gmail.com, cd722522@mail.ustc.edu.cn}\\
 $^{3}$Beijing Film Academy,  $^{4}$Microsoft Research,  $^{5}$Peking University \\
 \email{\{luyuan,ganghua\}@microsoft.com, ynh@ustc.edu.cn, baoquan@pku.edu.cn}
}

\maketitle

\begin{abstract}
Many different deep networks have been used to approximate, accelerate or improve traditional image operators, such as image smoothing, super-resolution and denoising. Among these traditional operators, many contain parameters which need to be tweaked to obtain the satisfactory results, which we refer to as ``parameterized image operators''. However, most existing deep networks trained for these operators are only designed for one specific parameter configuration, which does not meet the needs of real scenarios that usually require flexible parameters settings. To overcome this limitation, we propose a new decouple learning algorithm to learn from the operator parameters to dynamically adjust the weights of a deep network for image operators, denoted as the \textit{base} network. The learned algorithm is formed as another network, namely the \textit{weight learning} network, which can be end-to-end jointly trained with the \textit{base} network. Experiments demonstrate that the proposed framework can be successfully applied to many traditional parameterized image operators. We provide more analysis to better understand the proposed framework, which may inspire more promising research in this direction. Our codes and models have been released in \url{https://github.com/fqnchina/DecoupleLearning}.

\end{abstract} 

\section{Introduction}
Image operators are fundamental building blocks for many computer vision tasks, such as image smoothing \cite{fan2017generic,xu2011image}, super resolution \cite{kim2016accurate,ledig2017photo} and denoising \cite{mao2016image}. To obtain the desired results, many of these operators contain some parameters that need to be tweaked. We refer them as ``parameterized image operators'' in this paper. For example, parameters controlling the smoothness strength are widespread in most smoothing methods, and a parameter denoting the target upsampling scalar is always used in image super resolution.

Recently, many CNN based methods \cite{fan2017generic,kim2016accurate,xu2015deep} have been proposed to approximate, accelerate or improve these parameterized image operators  and achieved significant progress. However, we observe that the networks in these methods are often only trained for one specific parameter configuration, such as edge-preserving filtering \cite{fan2017generic} with a fixed smoothness strength, or super resolving low-quality images \cite{kim2016accurate} with a particular downsampling scale. Many different models need to be retrained for different parameter settings, which is both storage-consuming and time-consuming. It also prohibits these deep learning solutions from being applicable and extendable to a much broader corpus of images.

In fact, given a specific network structure, when training separated networks for different parameter configurations $\overrightarrow{\gamma}_k$ as \cite{fan2017generic, kim2016accurate, xu2015deep}, the learned weights $W_k$ are unconstrained and probably very different for each $\overrightarrow{\gamma}_k$. But can we find a common convolution weight space for different configurations by explicitly building their relationships? Namely, $W_k = h(\overrightarrow{\gamma}_k)$, where $h$ can be a linear or non-linear function. In this way, we can adaptively change the weights of the single target network based on $h$ in the runtime, thus enabling continuous parameter control.

To verify our hypothesis, we propose the first decouple learning framework for parameterized image operators by decoupling the weights from the target network structure. Specifically, we employ a simple \textit{weight learning} network $\mathcal{N}_{weight}$ as $h$ to directly learn the convolution weights of one task-oriented \textit{base} network $\mathcal{N}_{base}$.
These two networks can be trained end-to-end. During the runtime, the \textit{weight learning} network will dynamically update the weights of the \textit{base} network according to different input parameters, thus making the \textit{base} network generate different objective results. This should be a very useful feature in scenarios where users want to adjust and select the most visually pleasant results interactively.

We justify the effectiveness of the proposed framework for many different types of applications, such as edge-preserving image filtering with different degrees of smoothness, image super resolution with different scales of blurring, and image denoising with different magnitudes of noise. We also demonstrate the extensibility of our proposed framework on multiple input parameters for a specific application, and combination of multiple different image processing tasks. Experiments show that the proposed framework is able to learn as good results as the one solely trained with a single parameter value.

As an extra bonus, the proposed framework makes it easy to analyze the underlying working principle of the trained task-oriented network by visualizing different parameters. The knowledge gained from this analysis may inspire more promising research in this area. To sum up, the contributions of this paper lie in the following three aspects.

\begin{itemize}
\item We propose the first decouple learning framework for parameterized image operators, where a \textit{weight learning} network is learned to adaptively predict the weights for the task-oriented \textit{base} network in the runtime.
\item We show that the proposed framework can be learned to incorporate many different parameterized image operators and achieve very competitive performance with the one trained for a single specific parameter or operator.
\item We provide a unique perspective to understand the working principle of the trained task-oriented network with some valuable analysis and discussion, which may inspire more promising research in this area.
\end{itemize}

\section{Related Work}
In the past decades, many different image operators have been proposed for low level vision tasks. Previous work~\cite{xu2011image,karacan2013structure,xu2012structure,zhang2014rolling} proposed different priors to smooth images while preserving salient structures. Some work~\cite{buades2005non,elad2006image} utilized the spatial relationship and redundancy to remove unpleasant noise in the image. Some other papers~\cite{yang2010image,sun2008image,tipping2003bayesian} aimed to recover a high-resolution image from a low-resolution image. Among them, many  operators are allowed to tune some built-in parameters to obtain different results, which is the focus of this paper.

Recently, deep learning has been applied to many different tasks, like recognition \cite{zhang2018density,dai2017fason,Dai2017ICCV,cheng2017robust,cheng2018revisiting,li2017deep,zhang2018s3d}, generation \cite{qi2017global,li2018generating,lin2017adversarial}, and image to image translation \cite{fan2018revisiting,chen2017stylebank,chen2017coherent,chen2018stereoscopic,he2018deep,magan}. For the aforementioned image operators, some methods like \cite{fan2017generic,liu2016learning,xu2015deep} are also proposed to approximate, accelerate and improve them. 
But their common limitation is that one model can only handle one specific parameter. To enable all other parameters, enormous different models need to be retrained, which is both storage-consuming and time-consuming. 
By contrast, our proposed framework allows us to input continuous parameters to dynamically adjust the weights of the task-oriented \textit{base} network. Moreover, it can even be applied to multiple different parameterized operators with one single network.

Recently, Chen {\em et al.}~\cite{chen2017fast} conducted a naive extension for parameterized image operators by concatenating the parameters as extra input channels to the network. Compared to their method, where both the network structure and weights maintain the same for different  parameters, the weights of our \textit{base} network are adaptively changed. Experimentally we find our framework outperforms their strategy by integrating multiple image operators. By decoupling the network structure and weights, our proposed framework also makes it easier to analyze the underlying working principle of the trained task-oriented network, rather than leaving it as a black box as in many previous works like ~\cite{chen2017fast}.

Our method is also related to evolutionary computing and meta learning. Schmidhuber~\cite{schmidhuber1992learning} suggested the concept of fast weights in which one network can produce context-dependent weight changes for a second network. Some other works~\cite{Andrychowicz2016,Wichrowska2017a,Chen2017a} casted the design of an optimization algorithm as a learning problem, Recently, Ha {\em et al.}~\cite{ha2016hypernetworks} proposed to use a static hypernetwork to generate weights for a convolutional neural network on MNIST and Cifar classification. They also leverage a dynamic hypernetwork to generate weights of recurrent networks for a variety of sequence modelling tasks. The purpose of their paper is to exploit weight sharing property across different convolution layers. But in our cases, we pay more attention to the common shared property among numerous input parameters and many different image operators.

\section{Method}

\subsection{Problem Definition and Motivation}
The input color image and the target parameterized image operators are denoted as $\mathcal{I}$ and $f(\overrightarrow{\gamma}, \mathcal{I})$ respectively. $f(\overrightarrow{\gamma}, \mathcal{I})$  transforms the content of $\mathcal{I}$ locally or globally without changing its dimension. $\overrightarrow{\gamma}$ denotes the parameters which determine the transform degree of $f$ and may be a single value or a multi-value vector. For example, in $L_0$ smoothing\cite{l0smoothing2011}, $\overrightarrow{\gamma}$ is the balance weight controlling the smoothness strength, while in RTV filter \cite{xu2012structure}, it includes one more spatial gaussian variance. In most cases, $f$ is a highly nonlinear process and solved by iterative optimization methods, which is very slow in runtime.

Our goal is to implement parameterized operator $f$ with a base convolution network $\mathcal{N}_{base}$.  In previous methods like \cite{liu2016learning,xu2015deep}, given a specific network structure of $\mathcal{N}_{base}$, separated networks are trained for different parameter configuration $\overrightarrow{\gamma}_k$. In this way, the learned weights $\overrightarrow{W}_k$ of these separated networks are highly unconstrained and probably very different. But intuitively, for one specific image operator, the weights $\overrightarrow{W}_k$ of different $\overrightarrow{\gamma}_k$ might be related. So retraining separated models is too redundant. Motivated by this, we try to find a common weight space for different $\overrightarrow{\gamma}_k$ by adding a mapping constraint:  $\overrightarrow{W}_k = h(\overrightarrow{\gamma}_k)$, where $h$ can be a linear or non-linear function. 

In this paper, we directly learn $h$ with another \textit{weight learning} network $\mathcal{N}_{weight}$ rather than design it by handcraft. Assuming  $\mathcal{N}_{base}$ is a fully convolutional network having a total of $n$ convolution layers, we denote their weights as $\overrightarrow{W}_k=(W_1, W_2, ..., W_n)$ respectively, then

\begin{equation}
\begin{aligned}
(W_1, W_2, ..., W_n) = \mathcal{N}_{weight}(\overrightarrow{\gamma})
\end{aligned}
\end{equation}

where the input of $\mathcal{N}_{weight}$ is $\overrightarrow{\gamma}$ and the outputs are these weight matrices. In the training stage, $\mathcal{N}_{base}$ and $\mathcal{N}_{weight}$ can be jointly trained. In the inference stage, given different input parameter $\overrightarrow{\gamma}$, $\mathcal{N}_{weight}$ will adaptively change the weights of the target base network $\mathcal{N}_{base}$, thus enabling continuous parameter control.

Besides the original input image $\mathcal{I}$, the computed edge maps are shown to be a very important input signal for the target \emph{base} network in \cite{fan2017generic}. Therefore, we also pre-calculate the edge map $E$ of $\mathcal{I}$ and concatenate it to the original image as an extra input channel:
\begin{equation}
\begin{aligned}
E_{x,y} = \frac{1}{4}\sum_c(|\mathcal{I}_{x,y,c}-\mathcal{I}_{x-1,y,c}| + |\mathcal{I}_{x,y,c}-\mathcal{I}_{x+1,y,c}| \\ + |\mathcal{I}_{x,y,c}-\mathcal{I}_{x,y-1,c}| + |\mathcal{I}_{x,y,c}-\mathcal{I}_{x,y+1,c}|)
\end{aligned}
\end{equation}
where $x,y$ are the pixel coordinates and $c$ refers to the color channels.

To jointly train $\mathcal{N}_{base}$ and $\mathcal{N}_{weight}$, we simply use pixel-wise L2 loss in the RGB color space as \cite{chen2017fast} by default:
\begin{equation}
\begin{aligned}
\mathcal{L} = \lVert \mathcal{N}_{base}(\mathcal{N}_{weight}(\overrightarrow{\gamma}),\mathcal{I}, E) - f(\overrightarrow{\gamma}, \mathcal{I}) \rVert^2
\end{aligned}
\end{equation}

\begin{figure*}[t]
	\includegraphics[width=1.0\linewidth]{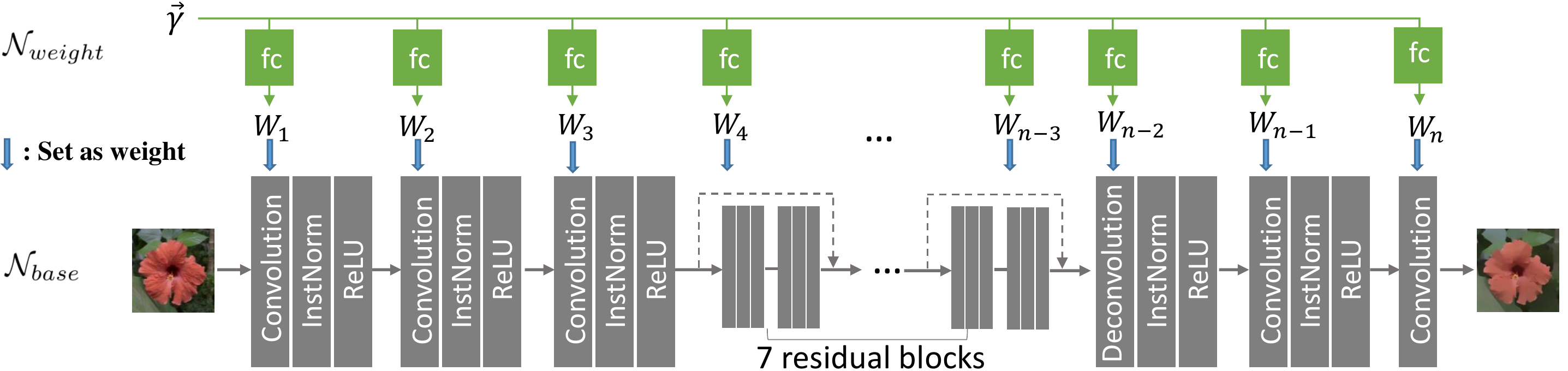}
	\caption{Our system consists of two networks: the above \textit{weight learning} network $\mathcal{N}_{weight}$ is designed to learn the convolution weights for the bottom \textit{base} network $\mathcal{N}_{base}$. Given a parameterized image operator constraint by $\protect\overrightarrow{\gamma}$,  these two networks are jointly trained, and $\mathcal{N}_{weight}$ will dynamically update the weights of $\mathcal{N}_{base}$ for different $\protect\overrightarrow{\gamma}$ in the inference stage.}
	\label{fg:arch}
\end{figure*}
\subsection{Network Structure}
As shown in \fref{fg:arch}, our \textit{base} network $\mathcal{N}_{base}$ follows a similar network structure as \cite{fan2017generic}. We employ 20 convolutional layers with the same $3\times3$ kernel size, among which the intermediate 14 layers are formed as residual blocks. Except the last convolution layer, all the former convolutional layers are followed by an instance normalization \cite{ulyanov2017improved} layer and a ReLU layer. To enlarge the receptive field of $\mathcal{N}_{base}$, the third convolution layer downsamples the dimension of feature maps by 1/2 using stride 2, and the third-to-last deconvolution layer (kernel size of $4\times4$) upsamples the downsampled feature maps to the original resolution symmetrically. In this way, the receptive field is effectively enlarged without losing too much image detail, and meanwhile the computation cost of intermediate layers is reduced. To further increase the receptive field, we also adopt dilated convolution \cite{yu2015multi} as \cite{chen2017fast}, more detailed network structure can be found in the supplementary material.

In this paper, the \textit{weight learning} network $\mathcal{N}_{weight}$ simply consists of 20 fully connected (fc) layers by default. The $i_{th}$ fc layer is responsible to learn the weights $W_i$ for the $i_{th}$ convolutional layer, which can be written as following:
\begin{equation}
\begin{aligned}
W_i = A_i\overrightarrow{\gamma} + B_i, \qquad\forall i \in \{1,2,...,20\}
\end{aligned}
\label{eq:fc_weight}
\end{equation}
Where $A_i, B_i$ are the weight and bias of the $i_{th}$ fc layer. Assuming the parameter $\overrightarrow{\gamma}$ has a dimension of $m$ and $W_i$ has a dimension of $n_{wi}$. The dimension of $A_i$ and $B_i$ would be $n_{wi}\times m$ and $n_{wi}$ respectively.

Note in this paper, we don't intend to design an optimal network structure neither for the \textit{base} network $\mathcal{N}_{base}$ nor the \textit{weight learning} network $\mathcal{N}_{weight}$. On the contrary, we care more about whether it is feasible to learn the relationship of the weights of $\mathcal{N}_{base}$ and different parameter configurations $\overrightarrow{\gamma}$ even by such a simple \textit{weight learning} network $\mathcal{N}_{weight}$.

\section{Experiments}

\subsection{Choice of Image Operators}
To evaluate the proposed framework on a broad scope of parameterized image operators, we leverage two representative types of image processing tasks: image filtering and image restoration. Within each of them, more than four popular operators are selected for detailed experiments.

\paragraph{\textbf{Image Filtering.}} Here we employ six popular image filters, denoted as $L_0$ \cite{xu2011image}, WLS \cite{farbman2008edge}, RTV \cite{xu2012structure}, RGF \cite{zhang2014rolling}, WMF \cite{zhang2014100} and shock filter \cite{osher1990feature}, which have been developed to work especially well for many different applications, such as image abstraction, detail exaggeration, texture removal and image enhancement. However, previous deep learning based approaches \cite{fan2017generic,liu2016learning,xu2015deep} are only able to deal with one single parameter value in one trained model, which is far from practical.

\paragraph{\textbf{Image Restoration.}} The goal of image restoration is to recover a clear image from a corrupted image. In this paper we deal with four representative tasks in this venue: super resolution \cite{dong2014learning,ledig2017photo}, denoising \cite{mao2016image,kligvasser2017xunit}, deblocking \cite{dong2015compression,tai2017memnet} and derain \cite{fu2017removing,zhang2018density}, which have been studied with deep learning based approaches extensively. For example, image super resolution is dedicated to increasing the resolution or enhancing the lost details from a low-resolution blurry image. To generate the pairwise training samples, previous work used to downsample a clear image by a specific scale with bicubic interpolation to synthesize a low-resolution image. Likewise, many previous works have typically been developed to fit a specific type of input image, such as a fixed upsampling scale.

\subsection{Implementation Details}

\paragraph{\textbf{Dataset.} } We take use of the 17k natural images in the PASCAL VOC dataset as the clear images to synthesize the ground truth training samples. The PASCAL VOC images are picked from Flicker, and consists of a wide range of viewing conditions. To evaluate our performance, 100 images from the dataset are randomly picked as the test data for the image filtering task. While for the restoration tasks, we take the well-known benchmark for each specific task for testing, which is specifically BSD100 (super resolution), BSD68 (denoise), LIVE1 (deblock), RAIN12 (derain). For the filtering task, we filter the natural images with the aforementioned algorithms to produce ground truth labels. As for the image restoration tasks, the clear natural image is taken as the target image while the synthesized corrupted image is used as input.

\paragraph{\textbf{Parameter Sampling.}} To make our network able to handle various parameters, we generate training image pairs with a much broader scope of parameter values rather than a single one. We uniformly sample parameters in either the logarithm or the linear space depending on the specific application. Regarding the case of logarithm space, let $l$ and $u$ be the lower bound and upper bound of the parameter, the parameters are sampled as follows:

\begin{equation} \label{equation:distribution}
y = e^x, \text{ where } x \in [{\ln l},{\ln u}]
\end{equation}

In other words, we first uniformly sample $x$ between $\ln l$ and $\ln u$, then map it back by the exponential function, similar to the one used in \cite{chen2017fast}. Note if the upper bound $u$ is tens or even hundreds of times larger than the lower bound $l$, the parameters are sampled in the logarithm space to balance their magnitudes, otherwise they are sampled in the linear space.

\setlength{\tabcolsep}{2pt}
\renewcommand{\arraystretch}{1}
\begin{table*}[t]
\begin{center}
\caption{Quantitative absolute difference between the network trained with a \textit{single} parameter value and \textit{numerous} random values for each image smoothing filter.}
\label{table:1}

\begin{tabular}{c  cccc  cccc  cccc }
\toprule[0.08em]
 & \multicolumn{4}{ c }{{$L_0$}} & \multicolumn{4}{ c }{{WLS}} & \multicolumn{4}{ c }{{RTV}} \\
\cmidrule{1-13}
 metric & $\lambda$  & single & nume. & diff &  $\lambda$  & single & nume. & diff  &  $\lambda$  & single & nume. & diff \\
\cmidrule{1-13}
\multirow{6}{*}{\small{PSNR}}
&0.002 & 40.69 & 39.46 & 1.23 & 0.100 & 44.00 & 42.12 & 1.88 & 0.002 & 41.11 & 40.66 & 0.45 \\
&0.004 & 38.96 & 38.72 & 0.24 & 0.215 & 43.14 & 42.64 & 0.50 & 0.004 & 40.91 & 41.10 & 0.19 \\
&0.020 & 36.07 & 35.71 & 0.36 & 1.000 & 41.93 & 41.63 & 0.30 & 0.010 & 40.50 & 41.07 & 0.57 \\
&0.093 & 33.08 & 31.92 & 1.16 & 4.641 & 39.42 & 39.64 & 0.22 & 0.022 & 41.07 & 40.77 & 0.30 \\
&0.200 & 31.75 & 30.43 & 1.32 & 10.00 & 39.13 & 38.51 & 0.62 & 0.050 & 40.73 & 39.18 & 1.55 \\
\cmidrule{2-13}
&ave.  & 36.11 & 35.25 & \textbf{0.86} & ave.  & 41.52 & 40.91 & \textbf{0.61} & ave.  & 40.86 & 40.55 & \textbf{0.31} \\
\cmidrule{1-13}
\multirow{6}{*}{\small{SSIM}}
&0.002 & 0.989 & 0.988 & 0.001 & 0.100 & 0.994 & 0.993 & 0.001 & 0.002 & 0.987 & 0.988 & 0.001 \\
&0.004 & 0.986 & 0.987 & 0.001 & 0.215 & 0.993 & 0.993 & 0     & 0.004 & 0.989 & 0.990 & 0.001 \\
&0.020 & 0.982 & 0.981 & 0.001 & 1.000 & 0.992 & 0.991 & 0.001 & 0.010 & 0.990 & 0.991 & 0.001 \\
&0.093 & 0.977 & 0.973 & 0.004 & 4.641 & 0.987 & 0.989 & 0.002 & 0.022 & 0.992 & 0.992 & 0 \\
&0.200 & 0.973 & 0.968 & 0.005 & 10.00 & 0.986 & 0.987 & 0.001 & 0.050 & 0.992 & 0.990 & 0.002 \\
\cmidrule{2-13}
&ave.  & 0.981 & 0.979 & \textbf{0.002} & ave.  & 0.990 & 0.990 & \textbf{0}     & ave.  & 0.990 & 0.990 & \textbf{0} \\
\bottomrule
\end{tabular}
\end{center}
\end{table*}

\subsection{Qualitative and Quantitative Comparison}
\textbf{Image Filtering. } We first experiment with our framework on five image filters. To evaluate the performance of our proposed algorithm, we train one network for each parameter value ($\lambda$) in one filter, and also train a network jointly on continuous random values sampled from the filter's parameter range, which can be inferred from the $\lambda$ column in Table \ref{table:1}. The performance of the two networks is evaluated on the test dataset with PSNR and SSIM error metrics. Since our goal is to measure the performance difference between these two strategies, we directly compute the absolute difference of their errors and demonstrate the results in Table \ref{table:1}. The results of the other two filters (RGF and WMF) are shown in the supplemental material due to space limitations.

As can be seen, though our proposed framework lags a little behind the one trained on a single parameter value, their difference is too small to be noticeable, especially for the SSIM error metric. Note that for each image filter, our algorithm only requires one jointly trained network, but previous methods need to train separate networks for each parameter value. Moreover even if the five filters are dedicated to different image processing applications, and varies a lot in their implementation details, our proposed framework is still able to learn all of them well, which verifies the versatility and robustness of our strategy.

\begin{figure*}[htp]
\begin{center}

\begin{tabular}{cc ccccc}

$\gamma$& input &0.002 & 0.004 & 0.020 & 0.093 & 0.200
\\

\raisebox{1.4cm}{\rotatebox[origin=c]{90}{\footnotesize{{$L_0$}}}}
&\includegraphics[width=1.90cm]{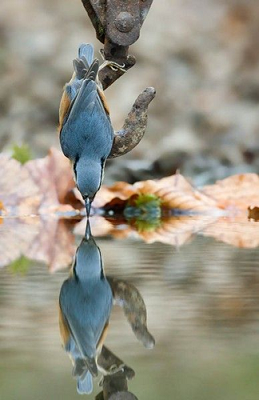}
&\includegraphics[width=1.90cm]{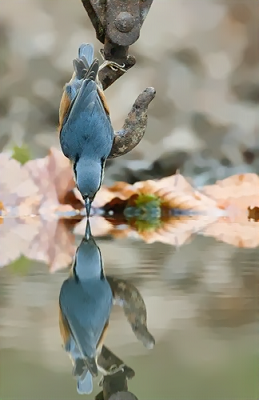}
&\includegraphics[width=1.90cm]{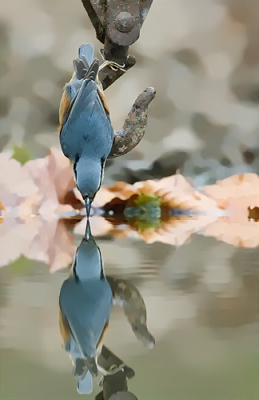}
&\includegraphics[width=1.90cm]{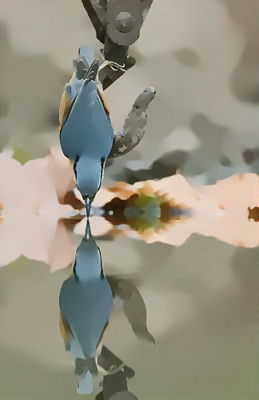}
&\includegraphics[width=1.90cm]{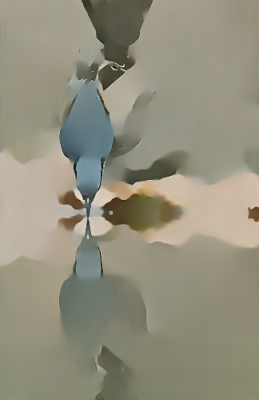}
&\includegraphics[width=1.90cm]{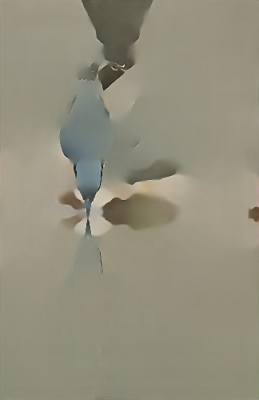}
\\

$\gamma$& input &0.002 & 0.04 & 0.010 & 0.022 & 0.050
\\

\raisebox{0.8cm}{\rotatebox[origin=c]{90}{\footnotesize{{RTV}}}}
&\includegraphics[width=1.90cm]{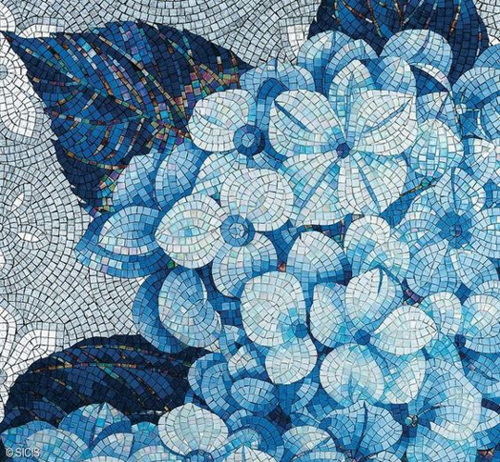}
&\includegraphics[width=1.90cm]{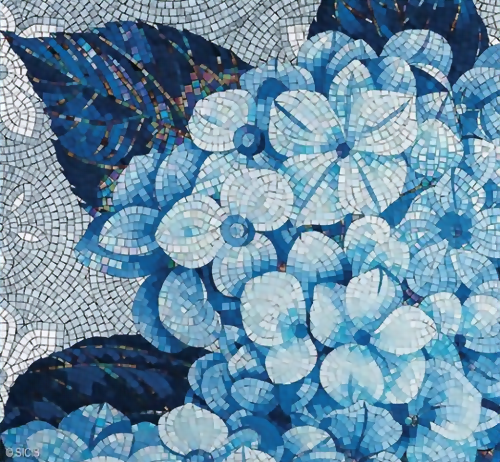}
&\includegraphics[width=1.90cm]{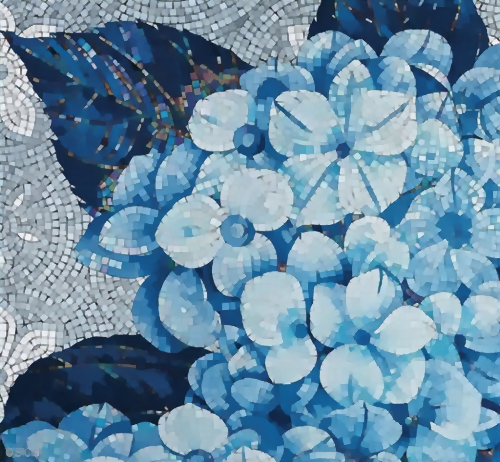}
&\includegraphics[width=1.90cm]{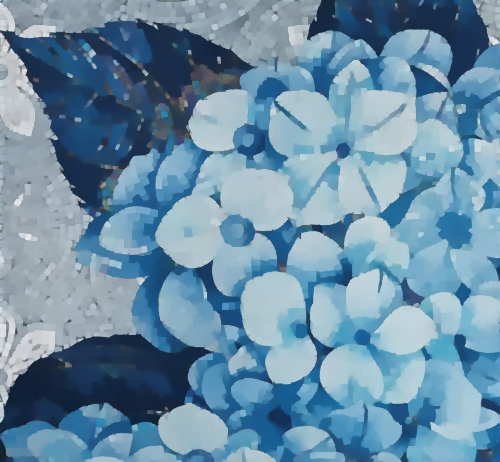}
&\includegraphics[width=1.90cm]{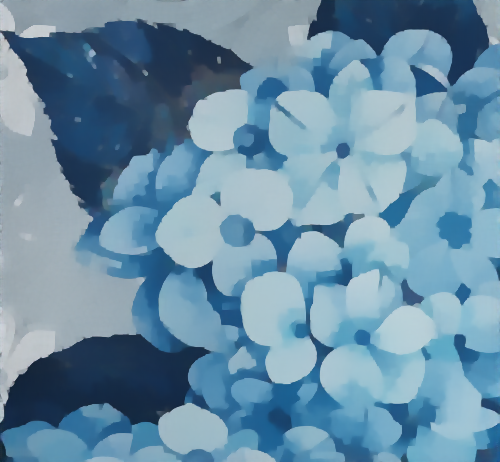}
&\includegraphics[width=1.90cm]{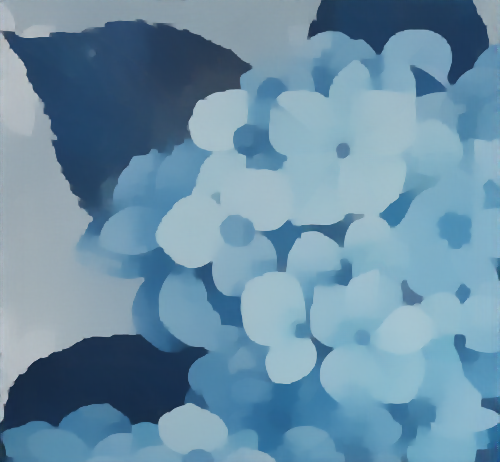}
\\

$\gamma$& input &1.00 & 3.25 & 5.50 & 7.75 & 10.00
\\

\raisebox{1.2cm}{\rotatebox[origin=c]{90}{\footnotesize{{RGF}}}}
&\includegraphics[width=1.90cm]{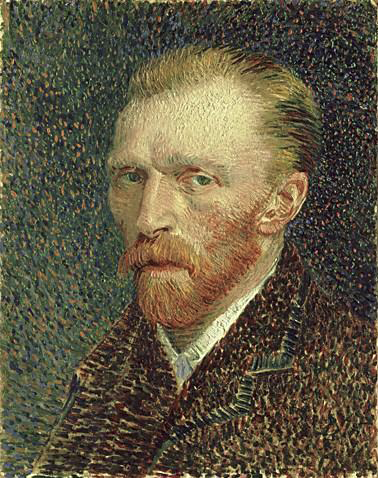}
&\includegraphics[width=1.90cm]{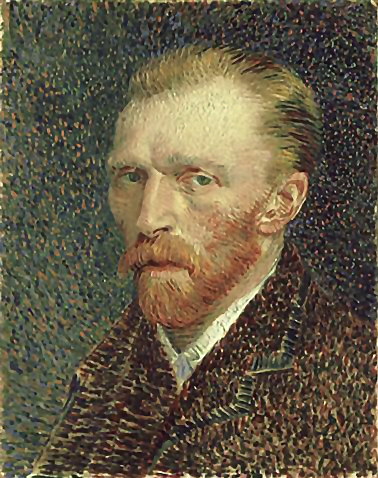}
&\includegraphics[width=1.90cm]{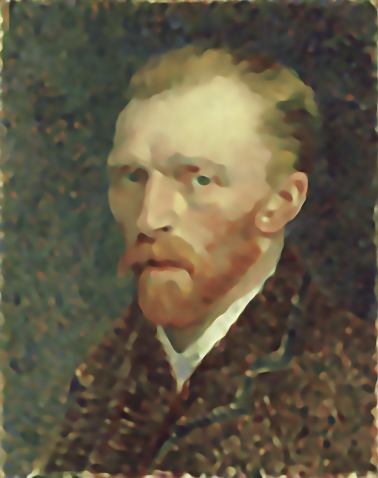}
&\includegraphics[width=1.90cm]{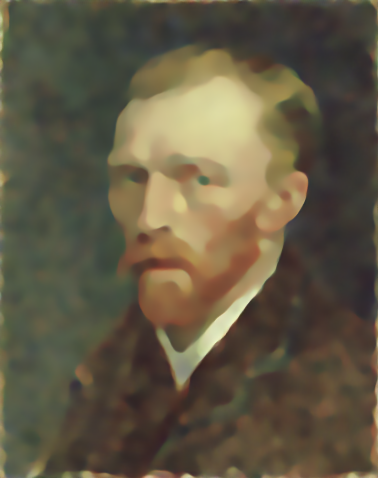}
&\includegraphics[width=1.90cm]{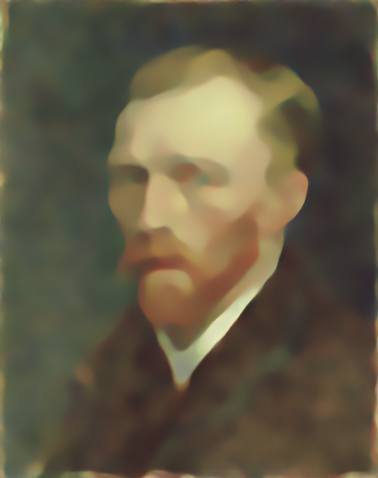}
&\includegraphics[width=1.90cm]{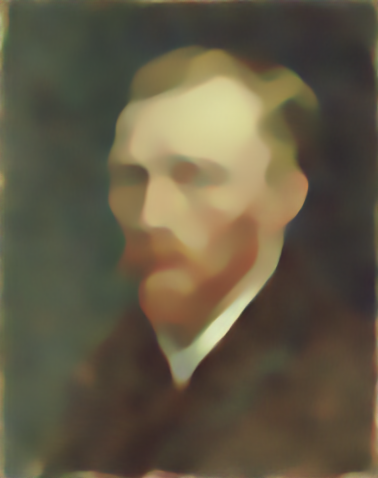}
\\

$\gamma$& input &1.00 & 3.25 & 5.50 & 7.75 & 10.00
\\

\raisebox{0.5cm}{\rotatebox[origin=c]{90}{\footnotesize{{WMF}}}}
&\includegraphics[width=1.90cm]{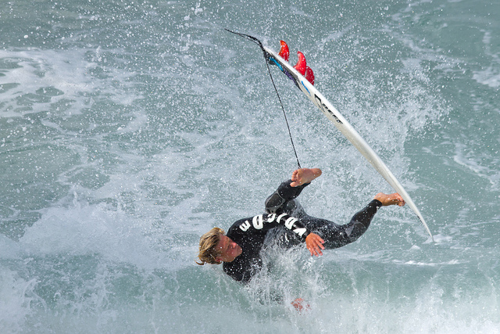}
&\includegraphics[width=1.90cm]{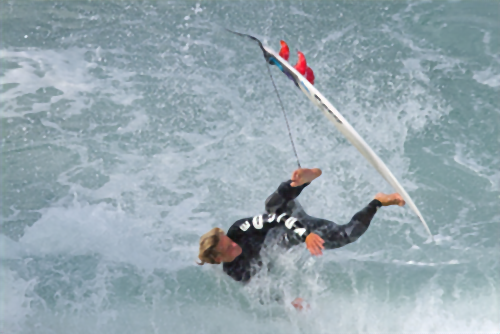}
&\includegraphics[width=1.90cm]{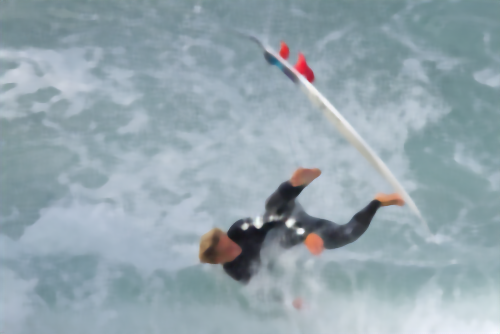}
&\includegraphics[width=1.90cm]{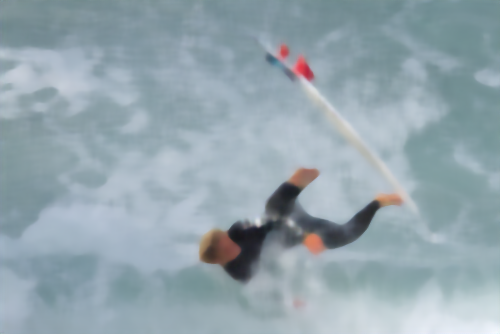}
&\includegraphics[width=1.90cm]{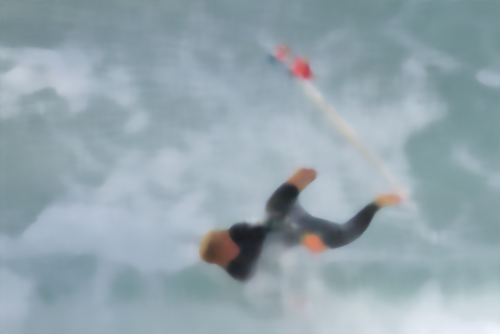}
&\includegraphics[width=1.90cm]{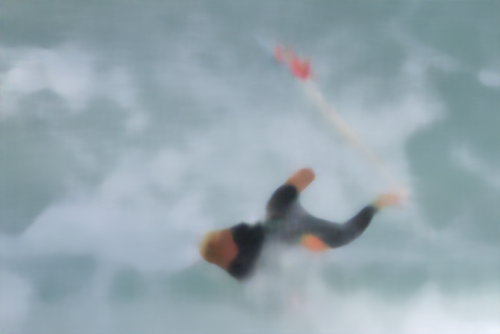}
\\

\end{tabular}

\end{center}
\vspace{-2mm}
\caption{Visual examples produced by our framework trained on continuous parameter settings of four image filters independently. Note all the smooth images for one filter are generated by a single network.}
\vspace{-2mm}
\label{figure:1}
\end{figure*}

Some visual results of our proposed framework are shown in Figure \ref{figure:1}. As can be seen, our single network trained on continuous random parameter values is capable of predicting high-quality smooth images of various strengths.

\setlength{\tabcolsep}{1pt}

\textbf{Image Restoration. } We then evaluate the proposed framework on three popular image restoration tasks as shown in Table \ref{table:2}, which perform essentially different from image filtering. Unlike the above operators which employ the filtered images as target to learn, this task takes the clear images as the ground truth label while the corrupted images as input. That is to say, as for the former task, given an input image, our network learns different filtering effects, while regarding the latter one, our model learns to recover from different corrupted images.

\setlength{\tabcolsep}{2pt}
\renewcommand{\arraystretch}{1}
\begin{table*}[t]
\begin{center}

\caption{Quantitative absolute difference in PSNR and SSIM between the network trained on a \textit{single} parameter value and \textit{numerous} random values on the three image restoration tasks. Their parameters specifically mean downsampling scale ($s$), Gaussian standard deviation ($\sigma$) and JPEG quality ($q$).}
\label{table:2}

\begin{tabular}{c  cccc  cccc  cccc }
\toprule[0.08em]
 & \multicolumn{4}{ c }{{Super Resolution}} & \multicolumn{4}{ c }{{Denoising}} & \multicolumn{4}{ c }{{Deblock}} \\
\cmidrule{1-13}
 metric & $s$  & single & nume. & diff &  $\sigma$  & single & nume. & diff  &  $q$  & single & nume. & diff \\
\cmidrule{1-13}
\multirow{4}{*}{\small{PSNR}}
& 2 & 31.78 & 31.62 & 0.16 & 15 & 31.17 & 31.07 & 0.10 & 10 & 29.26 & 29.17 & 0.09 \\
& 3 & 28.78 & 28.76 & 0.02 & 25 & 28.94 & 28.98 & 0.04 & 20 & 31.49 & 31.43 & 0.06 \\
& 4 & 27.31 & 27.31 & 0    & 50 & 26.22 & 26.14 & 0.08 &  \\
\cmidrule{2-13}
&ave.  & 29.29 & 29.23 & \textbf{0.06} & ave.  & 28.77 & 28.73 & \textbf{0.04} & ave.  & 30.37 & 30.30 & \textbf{0.07} \\
\cmidrule{1-13}
\multirow{4}{*}{\small{SSIM}}
& 2 & 0.894 & 0.892 & 0.002 & 15 & 0.881 & 0.883 & 0.002 & 10 & 0.817 & 0.817 & 0 \\
& 3 & 0.798 & 0.796 & 0.002 & 25 & 0.821 & 0.822 & 0.001 & 20 & 0.881 & 0.882 & 0.001 \\
& 4 & 0.728 & 0.726 & 0.002 & 50 & 0.722 & 0.718 & 0.004 & \\
\cmidrule{2-13}
&ave.  & 0.806 & 0.804 & \textbf{0.002} & ave.  & 0.808 & 0.807 & \textbf{0.001} & ave.  & 0.849 & 0.849 & \textbf{0} \\
\bottomrule
\end{tabular}
\end{center}
\end{table*} 

As shown in Table \ref{table:2}, our results trained jointly on continuous random parameter values also show no big difference from the one trained solely on an individual parameter value, which further validate our algorithm in a broader image processing literature.

\subsection{Extension to multiple input parameters}
Except for experimenting on a single input parameter, we also demonstrate our results on inputting multiple types of parameters, which is still very common for many image processing tasks.

In this section, we evaluate our performance on the famous texture removal tool RTV \cite{xu2012structure}. Likewise in previous experiments, we leverage $\lambda$ which balances between the data prior term and smoothness term in its energy function as one parameter, and $\sigma$ which controls the spatial scale for computing the windowed variation and is even more effective in removing textures. To generate the training samples, we randomly sample these two parameters. Therefore, the input parameter $\overrightarrow{\gamma}$ of the \textit{weight learning} network is a two-element vector $[\lambda, \sigma]$.

To evaluate the performance of our network on this two dimensional parameter space compared with the single parameter setting case, we sample a few parameters along one dimension while fixing another as shown in Table \ref{table:3}. We can see that for most of the 10 parameter settings, all achieve very close results to the one trained with an individual parameter setting. This verifies the effectiveness of our proposed network on this more difficult case.

\setlength{\tabcolsep}{5pt}
\renewcommand{\arraystretch}{1}
\begin{table*}[t]
\begin{center}
\caption{Quantitative comparison between the network trained on a \textit{single} parameter setting and \textit{numerous} random settings under the condition of multiple input parameters. Their absolute difference is shown besides the value of \textit{nume}. The results are tested by fixing one parameter while varying another.}
\label{table:3}
\begin{tabular}{  cccc  cccc }
\toprule[0.08em]
\multicolumn{4}{ c }{{RTV ($\lambda$ = 0.01)}} & \multicolumn{4}{ c }{{RTV ($\sigma$ = 3)}} \\
\cmidrule{1-8}
$\sigma$ & single & nume. & diff & $\lambda$ & single & nume. & diff \\
\cmidrule{1-8}
 2 & 40.53 & 40.39 & 0.14 & 0.002 & 41.11 & 40.17 & 0.94 \\
 3 & 39.52 & 40.76 & 1.24 & 0.004 & 40.91 & 40.78 & 0.13 \\
 4 & 41.19 & 41.06 & 0.13 & 0.010 & 40.50 & 40.76 & 0.26 \\
 5 & 41.29 & 41.26 & 0.03 & 0.022 & 41.07 & 40.45 & 0.62 \\
 6 & 41.81 & 41.19 & 0.62 & 0.050 & 40.73 & 38.52 & 2.21 \\
\cmidrule{1-8}
 ave & 40.86 & 40.93 & \textbf{\footnotesize{0.06}} & ave & 40.86 & 40.14 & \textbf{\footnotesize{0.72}} \\
\bottomrule
\end{tabular}
\end{center}
\end{table*}

\setlength{\tabcolsep}{2.5pt}
\renewcommand{\arraystretch}{1}
\begin{table*}[t]
\begin{center}
\caption{Numerical results (PSNR (above) and SSIM (bottom)) of our proposed framework jointly trained over different number of image operators (\#operators). ``6/4'' refers to the results jointly trained over either the front 6 filtering based approaches or the last 4 restoration tasks. ``10" is the results of jointly training all 10 tasks.}
\label{table:4}
\begin{tabular}{ c cccccccccc c }
\toprule[0.08em]
\cmidrule{1-12}
$\#$ope. & $L_0$ & WLS & RTV & RGF & WMF & shock & SR & denoise & deblock & derain & ave.\\
\cmidrule{1-12}
1    & 35.25 & 40.91 & 40.55 & 37.74 & 38.40 & 37.88 & 29.13 & 28.70 & 30.21 & 29.86 & \textbf{34.86} \\
6/4  & 33.54 & 38.02 & 37.69 & 35.90 & 36.46 & 35.27 & 28.89 & 28.67 & 30.10 & 30.32 & \textbf{33.49} \\
10   & 33.09 & 37.34 & 36.89 & 35.26 & 35.69 & 33.57 & 28.58 & 28.43 & 29.76 & 30.30 & \textbf{32.89} \\
\cmidrule{1-12}
1    & 0.979 & 0.991 & 0.990 & 0.984 & 0.980 & 0.987 & 0.804 & 0.804 & 0.847 & 0.893 & \textbf{0.925} \\
6/4  & 0.972 & 0.983 & 0.982 & 0.976 & 0.970 & 0.979 & 0.797 & 0.800 & 0.842 & 0.893 & \textbf{0.919} \\
10   & 0.967 & 0.980 & 0.978 & 0.973 & 0.966 & 0.970 & 0.791 & 0.792 & 0.838 & 0.890 & \textbf{0.914} \\
\bottomrule
\end{tabular}
\end{center}
\end{table*}

\subsection{Extension to joint training of multiple image operators} \label{sec:train_multi}

Intuitively, another challenging case for our proposed framework is to incorporate multiple distinct image operators into a single learned neural network, which is much harder to be trained due to their different implementation details and purposes. To explore the potential of our proposed neural network, we experiment by jointly training over (\textit{i}). 6 filtering based operators, (\textit{ii}). 4 image restoration operators or (\textit{iii}). all the 10 different operators altogether. To generate training images of each image operator, we sample random parameter values continuously within its parameter range. For the shock filter and derain task, we leverage its default parameter setting for training.

The input to the \textit{weight learning} network now takes two parameters, one indicates the specific image operator while the other is the random parameter values assigned to the specified filter. These 10 image operators are denoted simply by 10 discrete values that range from 0.1 to 1.0 in the input parameter vector. Since the absolute parameter range may differ a lot from operator to operator, for example, [2,4] for super resolution and [0.002,0.2] for $L_0$ filter, we rescale the parameters in all the operators into the same numerical range to enable consistent back-propagated gradient magnitude.

As shown in Table \ref{table:4}, training on each individual image operator achieves the highest numerical score (\#ope.=1), which is averaged over multiple different parameter settings just like in previous tables. While jointly training over either 6 image filters or 4 restoration tasks (\#ope.=6/4), even for the case where all 10 image operators are jointly trained (\#ope.=10), their average performance degrades but still achieves close results to the best score. It means with the same network structure, our framework is able to incorporate all these different image operators together into a single network without losing much accuracy.

Note that for the image restoration tasks, it is more meaningful not to specify parameters since in real life, users usually do not know the corruption degree of the input image. Therefore, we disable specifying parameters for the four restoration operators in this experiment. Surprisingly, we do not observe much performance degradation with this modification. Though it degrades the necessity of learning continuous parameter settings for image restoration tasks, it still makes a lot of sense by jointly training multiple image operators.

\subsection{Comparison with state-of-the-art image operators}

Note that we do not argue for the best performance in each specific task, since this is not the goal of this paper. Essentially, the performance on image operators is determined by the \textit{base} network structure, which is not our contribution, but many others \cite{fan2017generic,liu2016learning,xu2015deep} which develop more complex and advanced networks on each specific task. Even if this is not our goal, we still provide comparisons to demonstrate that our general framework performs comparably or even better than many previous work (one operator with one parameter).

Regarding \textit{image filtering}, the best performance is achieved by \cite{fan2017generic}. For the WLS filter example, with our simple and straightforward \textit{base} network trained with continuous parameter settings, we achieve very comparable results with \cite{fan2017generic} (PSNR/SSIM: 41.07/0.991 \emph{vs.} 41.39/0.994), which are superior to \cite{liu2016learning} (PSNR /SSIM: 38.29/0.983) and \cite{xu2015deep} (PSNR/SSIM: 33.92/0.963).

As for \textit{image restoration}, our framework trained with all four image restoration tasks performs better than DerainNet \cite{fu2017clearing} on the derain task (PSNR:30.32 vs 28.94 on RAIN12 dataset). And our model also achieves better PSNR (26.02) than many previous approaches BM3D \cite{dabov2007image} (25.62), EPLL \cite{zoran2011learning}(25.67), WNNM \cite{gu2014weighted} (25.87) on the BSD68 dataset for the denoising task.

\subsection{Understanding and analysis}
To better understand the \textit{base} network $\mathcal{N}_{base}$ and the \textit{weight learning} network $\mathcal{N}_{weight}$, we will conduct some analysis experiments in this section.

\paragraph{\textbf{The effective receptive field.}} In neuroscience, the receptive field is the particular region of the sensory space in which a stimulus will modify the firing of one specific neuron. The large receptive field is also known to be  important for modern convolutional networks. Different strategies are proposed to increase the receptive field, such as deeper network structure or dilated convolution. Though the theoretical receptive field of one network may be very large, the real effective receptive field may vary with different learning targets. So how is the effective receptive field of $\mathcal{N}_{base}$ changed with different parameters $\overrightarrow{\gamma}$ and $\mathcal{I}$ ? Here we use $L_0$ smoothing \cite{l0smoothing2011} as the default example operator.

\begin{figure*}
	\includegraphics[width=0.98\linewidth]{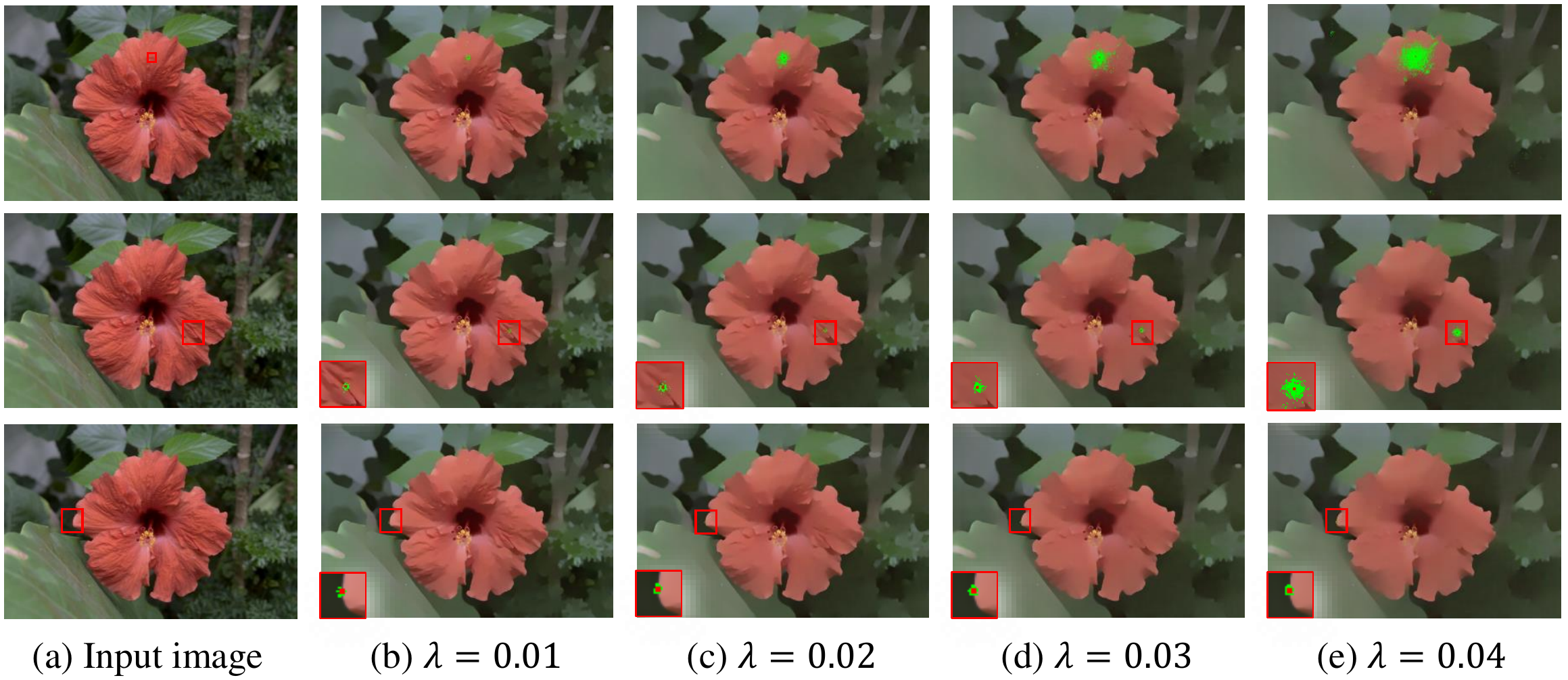}
\caption{Effective receptive field of $L_0$ smoothing for different spatial positions and parameter $\lambda$. The top to bottom indicate the effective receptive field of a non-edge point, a moderate edge point, and a strong edge point. }
\label{fg:receptive_field_study}

\end{figure*}

In \fref{fg:receptive_field_study}, we study the effective receptive field of a non-edge point, a moderate edge point, and a strong edge point with different smoothing parameters $\lambda$ respectively. To obtain the effective receptive field for a specific spatial point $p$, we first feed the input image into the network to get the smoothing result, then propagate the gradients back to the input while masking out the gradient of all points except $p$. Only the points whose gradient value is large than $0.025*grad_{max}$ ($grad_{max}$ is the maximum gradient value of input gradient) are considered within the receptive field and marked as green in \fref{fg:receptive_field_study}.
From \fref{fg:receptive_field_study}, we observe three important phenomena: 1) For a non-edge point, the larger the smoothing parameter $\lambda$ is, the larger the effective field is, and most effective points fall within the object boundary. 2) For a moderate edge point, its receptive field stays small until a relatively large smoothing parameter is used. 3) For a strong edge point, the effective receptive field is always small for all the different smoothing parameters. It means, on one hand, the \textit{weight learning} network $\mathcal{N}_{weight}$ can dynamically change the receptive field of $\mathcal{N}_{base}$ based on different smoothing parameters. On the other hand, the \textit{base} network $\mathcal{N}_{base}$ itself can also adaptively change its receptive field for different spatial points.

\paragraph{\textbf{Decomposition of the weight learning network}}

To help understand the connection between the \textit{base} network $\mathcal{N}_{base}$ and the \textit{weight learning} network $\mathcal{N}_{weight}$, we decompose the parameter vector $\overrightarrow{\gamma}$ and the weight matrix $A_i$ into independent elements $\gamma_1,...,\gamma_m$ and $A_{i0}, ...,A_{im}$ respectively, then:

\begin{equation}
\begin{aligned}
(A_i\overrightarrow{\gamma} + B_i)\otimes x = \sum_{k=1}^m \gamma_k A_{ik} \otimes x + B_i\otimes x
\end{aligned}
\label{fg:fc_analysis}
\end{equation}

where $\otimes$ denotes convolution operation, and $m$ is the dimension of $\overrightarrow{\gamma}$. In other words, the one convolution layer, whose weights are learned with one single fc layer, is exactly equivalent to a multi-path convolution block as shown in \fref{fg:fc_analysis}. Learning the weight and bias of the single fc layer is equivalent to learning the common basic convolution kernels $B_i, A_{i1}, A_{i2},...,A_{im}$ in the convolution block.

\begin{figure*}[t]
\includegraphics[width=0.9\linewidth]{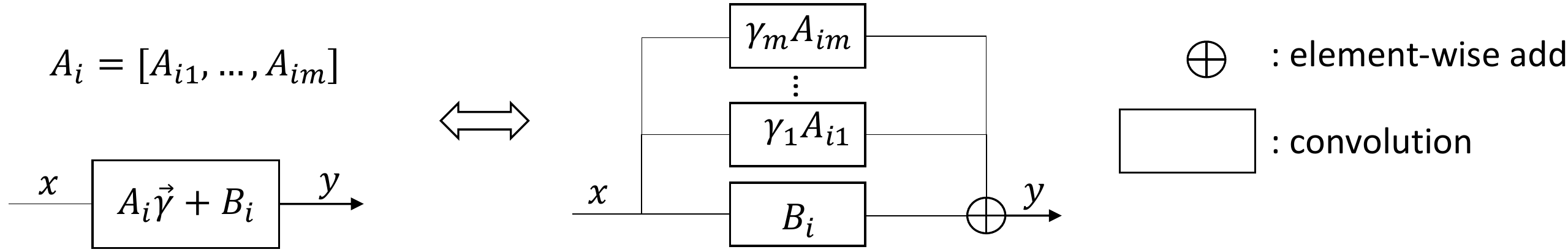}
\caption{Equivalent analysis of the connection between the \textit{base} network $\mathcal{N}_{base}$ and the \textit{weight learning} network $\mathcal{N}_{weight}$. One convolution layer whose weights are learnt by the fc layer is exactly equivalent to a multi-path convolution blocks.}
\label{fg:fc_analysis}

\end{figure*}

\paragraph{\textbf{Visualization of the learned convolution weights}}

The learned convolution weights can be generally classified into two classes: kernels generated by different parameter values of a single image operator, and kernels generated by different image operators. We analyse both groups of kernels on the model trained on 10 image operators which is introduced in subsection \ref{sec:train_multi}. In this case, the input to the weight learning network takes two parameters, hence the learned convolution weights for a specific layer $i$ in the base network should be,

\begin{align}
\begin{split}\label{eq:1}
    W_i {} & = \gamma_{1} A_{i1} + \gamma_{2} A_{i2} + B_i
\end{split}
\end{align}

where $\gamma_{1}$ refers to the input parameter value of a specific operator, and $\gamma_{2}$  indicates the type of the operator, which is defined by ten discrete numbers that range from ``0.1'' to ``1.0'' for different operators separately. $A_{i1}$ and $A_{i2}$ are its corresponding weights in the fully connected layer. Therefore, for a single image operator, $\gamma_{2} A_{i2} + B_i$ is a fixed value and the only modification to its different parameter values is $\gamma_{1} A_{i1}$, which scales a high-dimension value. That is to say, each time when one adjusts the operator parameter by $\gamma_{1}$, the learned convolution weights are only shifted to some extent in a fixed high-dimensional direction. Similar analysis also applies to the transformation of different operators.

\begin{figure*}[t]
\begin{center}
\begin{tabular}{c}
\includegraphics[width=6cm]{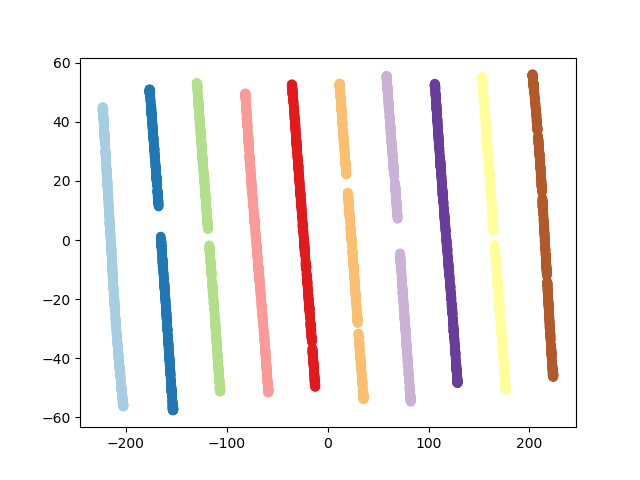}
\end{tabular}
\end{center}

\caption{T-SNE illustration of the learned weights of the 2nd convolution layer in the base network. The displayed convolution weights are generated by the jointly trained network with 10 image operators. Each color indicates one specific operator. We also observe similar visualized results for the other convolution layers.}

\label{figure:2_supp}
\end{figure*} 

We visualize the learned convolution kernels via t-SNE in Figure \ref{figure:2_supp}. Each color indicates one image operator, and for each operator, we randomly generate 500 groups of convolution weights with different parameters. As can be seen, the distance of every two adjacent operator is almost the same, it shifts along the x dimension for a fixed distance. For a single filter, while adjusting the parameters continuously, the convolution weights shift along the y dimension. This figure just conforms to our analysis about the convolution weights in the high-dimensional space. It is very surprising that all different kinds of learned convolution weights can be related with a high-dimensional vector, and the transformation between them can be represented by a very simple linear function.

As analyzed in the supplemental material, the solution space of an image processing task could be huge in the form of learned convolution kernels. Two exactly same results may be represented by very different convolution weights. The linear transformation in our proposed weight learning network actually connects all the different image operators and constrains their learned convolution weights in a limited high dimensional space.

\section{Conclusion}
In this paper, we propose the first decouple learning framework for parameterized image operators, where the weights of the task-oriented \textit{base} network $\mathcal{N}_{base}$ are decoupled from the network structure and  directly learned by another \textit{weight learning} network $\mathcal{N}_{weight}$. These two networks can be easily end-to-end trained, and $\mathcal{N}_{weight}$ dynamically adjusts the weights of $\mathcal{N}_{base}$ for different parameters $\overrightarrow{\gamma}$ during the runtime. We show that the proposed framework can be applied to different parameterized image operators, such as image smoothing, denoising and super resolution, while obtaining comparable performance as the network trained for one specific parameter configuration. It also has the potential to jointly learn multiple different parameterized image operators within one single network. To better understand the working principle, we also provide some valuable analysis and discussions, which may inspire more promising research in this direction. More theoretical analysis is worthy of further exploration in the future. 

\noindent\textbf{Acknowledgement} \small{This work was supported in part by: National 973 Program (2015CB352501), NSFC-ISF (61561146397), the Natural Science Foundation of China under Grant U1636201 and 61629301}

\bibliographystyle{splncs04}
\bibliography{egbib}

\end{document}